\documentclass[10pt,twocolumn,letterpaper]{article}

\usepackage{iccv}
\usepackage{times}
\usepackage{epsfig}
\usepackage{graphicx}
\usepackage{amsmath}
\usepackage{amssymb}
\usepackage{bm}
\usepackage{multirow}
\usepackage[tight, footnotesize]{subfigure}
\usepackage{array}
\usepackage{paralist}
\usepackage{enumitem}
\usepackage{color}
\usepackage{booktabs}
\usepackage[pagebackref, breaklinks, letterpaper, colorlinks, linkcolor=blue, citecolor=blue]{hyperref}

\newcommand{\bx}{\mathbf{x}}

\usepackage{color}

\definecolor{gray}{rgb}{0.35,0.35,0.35}
\definecolor{MyBlue}{rgb}{0,0.2,0.8}
\definecolor{MyRed}{rgb}{0.8,0.2,0}
\definecolor{MyGreen}{rgb}{0.0,0.5,0.1}
\definecolor{MyGray}{rgb}{0.4,0.4,0.4}

\long\def\ignorethis#1{}
\def\etal{et~al.\xspace}

\ignorethis{
\newcommand*\patchAmsMathEnvironmentForLineno[1]{%
  \expandafter\let\csname old#1\expandafter\endcsname\csname #1\endcsname
  \expandafter\let\csname oldend#1\expandafter\endcsname\csname end#1\endcsname
  \renewenvironment{#1}%
     {\linenomath\csname old#1\endcsname}%
     {\csname oldend#1\endcsname\endlinenomath}}%
\newcommand*\patchBothAmsMathEnvironmentsForLineno[1]{%
  \patchAmsMathEnvironmentForLineno{#1}%
  \patchAmsMathEnvironmentForLineno{#1*}}%
\AtBeginDocument{%
\patchBothAmsMathEnvironmentsForLineno{equation}%
\patchBothAmsMathEnvironmentsForLineno{align}%
\patchBothAmsMathEnvironmentsForLineno{flalign}%
\patchBothAmsMathEnvironmentsForLineno{alignat}%
\patchBothAmsMathEnvironmentsForLineno{gather}%
\patchBothAmsMathEnvironmentsForLineno{multline}%
} 
}

\iccvfinalcopy 

\begin{document}

\title{Learning Structured Semantic Embeddings for Visual Recognition}

\author{
Dong Li$^1$, Hsin-Ying Lee$^3$, Jia-Bin Huang$^2$, Shengjin Wang$^1$, and Ming-Hsuan Yang$^3$ \\
$^1$Tsinghua University, ~$^2$Virginia Tech, ~$^3$University of California, Merced
}

\maketitle


\begin{abstract}
Numerous embedding models have been recently explored to incorporate semantic knowledge into visual recognition. 
Existing methods typically focus on minimizing the distance between the corresponding images and texts in the embedding space but do not explicitly optimize the underlying structure.
Our key observation is that modeling the pairwise image-image relationship improves the discrimination ability of the embedding model.
In this paper, we propose the structured discriminative and difference constraints to learn visual-semantic embeddings.
First, we exploit the discriminative constraints to capture the intra- and inter-class relationships of image embeddings.
The discriminative constraints encourage separability for image instances of different classes.
Second, we align the difference vector between a pair of image embeddings with that of the corresponding word embeddings. 
The difference constraints help regularize image embeddings to preserve the semantic relationships among word embeddings. 
Extensive evaluations demonstrate the effectiveness of the proposed structured embeddings for single-label classification, multi-label classification, and zero-shot recognition.
\end{abstract}

\section{Introduction}
\label{section: introduction}

%
Recent visual recognition methods typically train multi-class classifiers using image datasets labeled with a pre-defined set of \emph{discrete} classes~\cite{krizhevsky2012imagenet,simonyan2014very,szegedy2014going}.
However, such classifiers are not capable of capturing semantic relationships among visual categories since they are trained in the discrete label space.
For example, discrete classifiers treat the three classes \emph{cat}, \emph{dog} and \emph{bicycle} as unrelated and distinct categories. 
As a result, they cannot encode the fact that the two classes \emph{cat} and \emph{dog} are semantically more similar than that between \emph{cat} and \emph{bicycle}.
Furthermore, to recognize a new category, the discrete classifiers need to be re-trained on a sufficient amount of training examples of the new class.
The lack of semantic information transfer substantially limits the visual recognition methods to scale up to large numbers of classes.

To address these issues, visual-semantic embedding models~\cite{frome2013devise,norouzi2013zero,socher2013zero} have been proposed to leverage the semantic knowledge from text data.
By using a large set of unannotated text data, we can construct a \emph{continuous} and \emph{semantically meaningful} word embedding space~\cite{mikolov2013distributed}.
Images can then be mapped into the same semantic space to align the embeddings of their corresponding labels (typically by minimizing ranking losses).
The rich semantic relationships from the text data help better recognize visual categories, reduce semantically implausible predictions, and enable zero-shot recognition.
%

\begin{figure}[t]
\centering
\includegraphics[width=\linewidth]{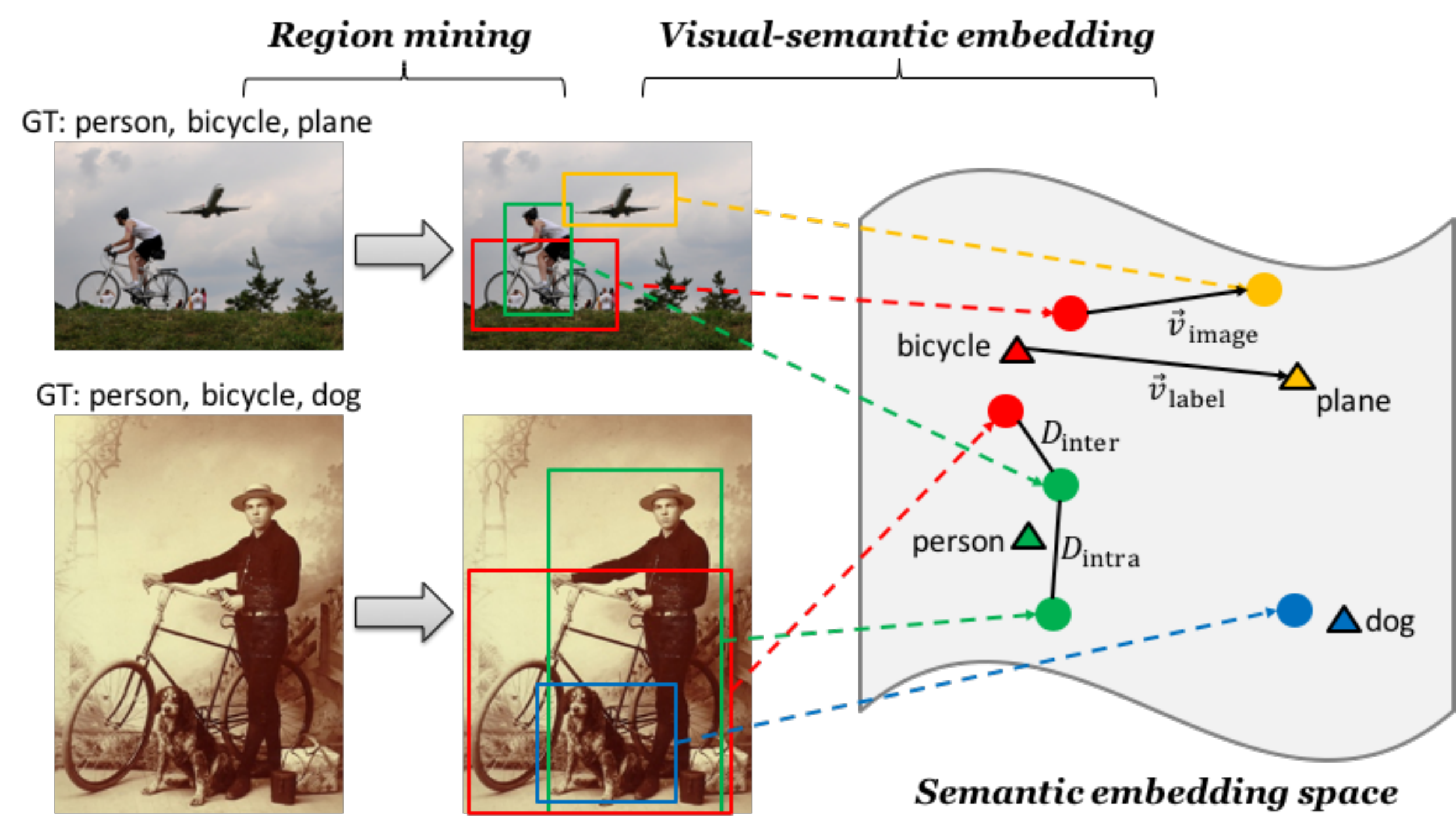}
\caption{
Illustration of the proposed constraints for learning visual-semantic embeddings.
Triangles represent label (word) embeddings, and circles represent image embeddings.
The visual categories are color-coded. 
(1) Discriminative constraints (Section~\ref{section: discriminative}) capture the intra- and inter-class relationships of image embeddings (\eg, $D_\mathrm{intra} < D_\mathrm{inter}$). 
(2) Difference constraints (Section~\ref{section: difference}) align the difference vector between a pair of image embeddings with that of the corresponding label embeddings (\eg, $\vec{v}_\mathrm{image}$ and $\vec{v}_\mathrm{label}$ should be as similar as possible).
}
\label{figure: overview}
\vspace{-2mm}
\end{figure}

Much effort has been made to learn visual-semantic embeddings by diverse semantic knowledge sources~\cite{hwang2013analogy,hwang2014unified,wang2015learning,vendrov2015order}. 
For example, analogy-preserving embeddings~\cite{hwang2013analogy} use analogical parallelogram constraints to reflect the relationships between multiple pairs of classes.
The analogy relationships can help disambiguate the semantically similar categories. 
However, learning analogy-preserving embeddings requires manual annotations of attributes and off-the-shelf classifiers to discover a set of analogies. 
This limits the scalability of the embedding model to handle large numbers of object categories. 
Structure-preserving constraints are also explored to model the neighborhood structure within each modality~\cite{wang2015learning}. 
Such constraints help improve image-to-text or text-to-image matching by reducing the distance between semantically similar instances. 
However, they model the neighborhood structure for images and texts separately with two independent regularization terms and thus cannot preserve the semantic relationships between a pair of word embeddings.

In this paper, we propose to learn visual-semantic embeddings by incorporating \emph{discriminative} and \emph{difference} constraints as shown in Figure~\ref{figure: overview}.
We exploit the discriminative constraints to explicitly model the intra- and inter-class relationships of image embeddings. 
Specifically, we explore two types of discriminative constraints (contrastive loss and triplet loss), both of which can help improve the discrimination ability of the embedding model. 
%
%
%
While discriminative constraints encourage separability for image instances of different categories, there are no constraints on \emph{how} the two instances should be pulled apart.
To alleviate these ambiguity issues, we propose the difference constraints to regularize the learning of visual-semantic embedding model.
The difference constraints enforce two image embeddings to have similar relative positions with their corresponding label embeddings.
Similar to recent work~\cite{ren2015multi,karpathy2014deep,karpathy2015deep}, we 
extend the embedding model to address the \emph{multi-label} scenario where each image may contain multiple labels. 
%
%
Through extensive evaluations, we demonstrate the effectiveness of the proposed structured embeddings for visual recognition. 
For recognizing seen classes, our method performs better over baseline methods on the CIFAR-10/100 datasets for single-label image classification and achieves competitive performance with the state-of-the-arts on the NUS-WIDE dataset for multi-label image classification.
For recognizing unseen classes, our method performs favorably against the state-of-the-arts on the aP\&Y and large-scale ImageNet datasets.

We make the following contributions in this work:

First, we exploit the discriminative constraints to learn image-text embeddings using a multi-task learning strategy. The discriminative constraints explicitly model the intra- and inter-class relationships. We show that two types of discriminative constraints can help improve the discrimination ability of the embedding model.
%

Second, we propose the difference constraints for aligning the difference vectors of image pairs with those of the corresponding label pairs.
The difference constraints serve as a regularizer to help learn image embeddings with proper semantic relationships among the various categories.
%

Third, we present a unified learning formulation that learns visual-semantic embeddings with two additional structured constraints while drawing relations between them. 
Extensive experimental results show that learning with the two complementary structured constraints significantly 
improves visual recognition tasks, including single-label classification, multi-label classification, and zero-shot recognition.
%


\section{Related Work}
\label{section: related-work}


\paragraph{Visual-semantic embedding.}
Visual-semantic embedding models relate information from different domains, such as images and texts.
%
%
Attributes can be used to capture semantic properties shared across different classes~\cite{farhadi2009describing,lampert2014attribute}.
However, attribute-based approaches do not scale up to large amounts of categories due to manually defined attribute ontology and expensive labeling effort.
Another line of work leverages neural language models to incorporate semantic knowledge for learning image embeddings~\cite{frome2013devise,norouzi2013zero,ren2015multi,wang2015learning,hwang2013analogy,hwang2014unified}. 
In these approaches, the language model learns semantically meaningful word embeddings from unannotated text data (\eg,~\cite{mikolov2013distributed}).
Ranking losses~\cite{frome2013devise,ren2016joint} are typically used to learn the image embedding space by constraining the distance between the image embedding and the corresponding word embedding smaller than that between the image embedding and other randomly chosen words. Images are thus projected to nearby positions with their corresponding labels in the semantic space.

\vspace{-3mm}
\paragraph{Learning embedding with constraints.}
Other semantic knowledge has also been used to improve embedding models.
Examples include analogies~\cite{hwang2013analogy}, taxonomies~\cite{hwang2014unified}, hierarchies~\cite{vendrov2015order} and neighborhood structures~\cite{wang2015learning}. 
Our work is related to~\cite{wang2015learning} in the aspect of modeling the neighborhood structure of image embeddings. 
In contrast to learning the transformation layers only, we train our entire network for improved adaptation of visual representations. 
Moreover, unlike the constraints in~\cite{wang2015learning} that preserve the local neighborhood structure for images and texts \emph{separately}, we regularize that pairs of image embeddings have similar relative positions with their corresponding label embeddings.
The proposed difference constraints bear some resemblance with the analogical parallelogram constraints~\cite{hwang2013analogy}, but differ in three aspects. 
First, we do not rely on costly attribute annotations and off-the-shelf classifiers to discover analogies. 
Second, the analogy constraints use one pair of classes to help recognize another pair.
In contrast, we align the difference vectors between images and labels from the \emph{same} pair of two classes.
Third, we incorporate both discriminative and difference constraints into a unified deep learning framework.
The contrastive loss~\cite{chopra2005learning,zhang2016tracking} or triplet loss~\cite{schroff2015facenet,zhang2016embedding,wang2015unsupervised} has been  
applied for feature learning.
In the context of visual-semantic embedding, we apply either of them as discriminative constraints to improve the embedding baseline using a multi-task learning strategy.

Language grounding methods~\cite{kiros2014multimodal,Kottur2015visual,lazaridou2015combining} have been recently proposed for cross-modal tasks (\eg, image-to-text and text-to-image retrieval) by jointly optimizing visual and semantic embeddings.
In this work, we focus on improving the visual model given the pre-trained semantic model.

\vspace{-3mm}
\paragraph{Convolutional neural networks for visual recognition.}
Convolutional neural networks (CNNs) have shown promising results on various visual recognition tasks, \eg, image classification~\cite{krizhevsky2012imagenet,simonyan2014very,szegedy2014going}.
Recent work addresses the multi-label recognition problem in the discrete label space~\cite{gong2013deep,Wang-CVPR-2016,Li-CVPR-2016}.
For learning visual-semantic embeddings in the general multi-label settings, existing methods either use multiple instance learning~\cite{ren2015multi} or apply off-the-shelf detectors~\cite{karpathy2014deep,karpathy2015deep} to generate candidate regions for each label. 
We adopt a different strategy to mine associated regions for each label via a multi-label training procedure. 
The model pre-trained on the multi-label classification task is also used as an initialization for the subsequent learning steps.

\vspace{-3mm}
\paragraph{Zero-shot learning.}
The goal of zero-shot learning is to recognize \emph{unseen} classes without any training. 
As visual examples of test classes are not available during the training process, auxiliary sources are required to relate the unseen classes with the seen classes.
The semantic information sharing across categories can be achieved by attributes~\cite{farhadi2009describing,romera2015embarrassingly,lampert2014attribute,bucher2016improving}, word embeddings~\cite{frome2013devise,norouzi2013zero,socher2013zero}, or a combination of multiple semantic sources~\cite{fu2015zero,akata2015evaluation}.
Prior work addresses this problem by learning attribute classifiers~\cite{lampert2014attribute,rohrbach2011evaluating} or compatibility functions~\cite{XASNHS16,frome2013devise,romera2015embarrassingly,bucher2016improving,socher2013zero}.
%
Recent methods also consider generalized zero-shot learning where test data may come from seen classes and the label space is the union of both seen and unseen classes~\cite{chao2016empirical,XianCVPR2017}.
%


\section{Approach}
\label{section: method}

Our goal is to learn structured semantic embeddings for visual recognition.
We build our learning framework based on deep convolutional neural networks. 
The pre-trained word embedding model provides continuous vector representations of each image label for training the CNN.
To address the multi-label case, we first train the network for multi-label image classification. 
The learned CNN model is then used to mine top candidate image regions for each label (Section~\ref{section: mining}). 
Using the mined regions as training instances, we retrain the network to embed image features to the semantic embedding space with our structured constraints (Section~\ref{section: embedding}).
We include implementation details in Section~\ref{section: details}.

\begin{figure}[t]
\centering
\includegraphics[width=\linewidth]{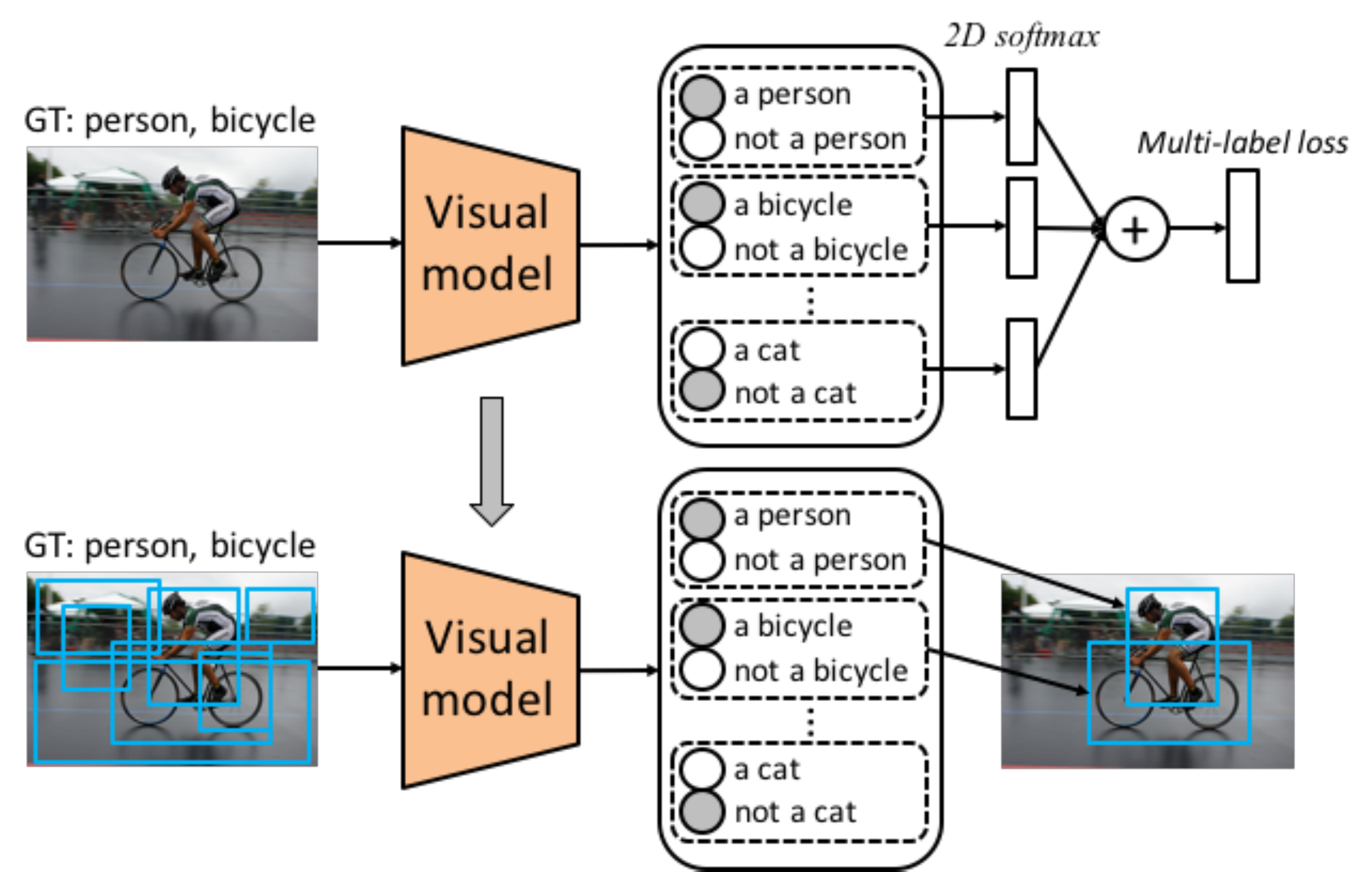}
\caption{Region mining via multi-label training. First, we use the multi-label loss in~\cite{Li-CVPR-2016} to train the network for multi-label image classification. Second, we compute the classification scores of candidate region proposals and select the best-matched region with the highest score for each ground-truth (GT) image label.
}
\vspace{-2mm}
\label{figure: mining}
\end{figure}

\subsection{Region Mining}
\label{section: mining}

Existing visual-semantic embedding methods typically address the single-label setting where each image contains only one semantic label (\eg, ImageNet).
This substantially limits the applicability of the learned embedding models for visual recognition as real-world images often contain multiple labels. 
Similar to recent work~\cite{ren2015multi,karpathy2014deep,karpathy2015deep}, we address this problem by assigning labels to image regions.
To this end, we use the multi-label loss~\cite{Li-CVPR-2016} to train the network for multi-label image classification.
For each ground-truth image label, we compute the classification scores of general region proposals~\cite{krahenbuhl2014geodesic} 
and select the best-matched region with the highest score.
Figure~\ref{figure: mining} illustrates the multi-label classification training process for mining regions that correspond to the image-level labels.

%

\subsection{Structured Semantic Embedding}
\label{section: embedding}

We use the mined regions as training instances to learn the visual embedding model. 
We denote the training set as $\mathcal{D} = \{(\bx_i, y_i)\}_{i=1}^N$, where $\bx_i$ indicates the $i_{th}$ region instance and $y_i$ indicates the corresponding label.
Our goal is to learn a mapping function $f_\Theta$ that maps from the image space $\mathcal{I}$ to a continuous semantic space $\mathcal{S}$, $f_\Theta: \mathcal{I} \to \mathcal{S}$, where $\Theta$ denotes the network parameters to be optimized.
%
%
For the word embedding, we exploit the pre-trained word2vec model~\cite{mikolov2013distributed} on the Google News dataset ($\sim$100 billion words) to generate a 300D vector representation for each label. We denote $s(\cdot)$ as the label embedding function learned by the word2vec model.
For the image embedding, we train the CNN to learn $\Theta$ by mapping an image to the same 300D space~$\mathcal{S}$. For simplicity, we denote $f(\cdot)$ instead of $f_\Theta(\cdot)$ as the image embedding function. Both image and label embeddings are normalized to unit norm.

\begin{figure}[t]
\footnotesize
\begin{center}
\begin{tabular}{@{}cc@{}}
\includegraphics[width = 0.4\linewidth]{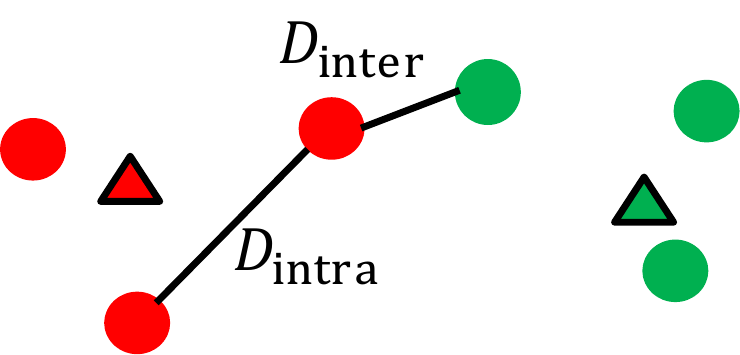} & 
\includegraphics[width = 0.4\linewidth]{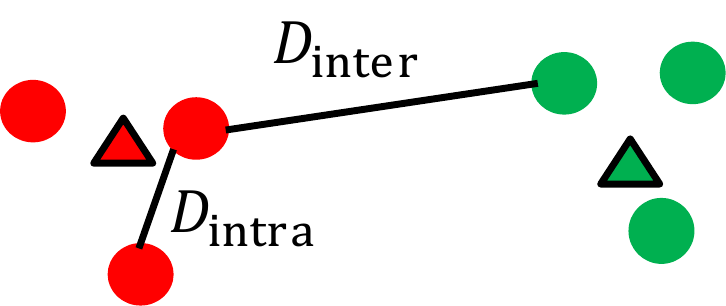} \\ 
(a) w/o discriminative constraints & 
(b) w/ discriminative constraints \vspace{+2mm} \\
\includegraphics[width = 0.4\linewidth]{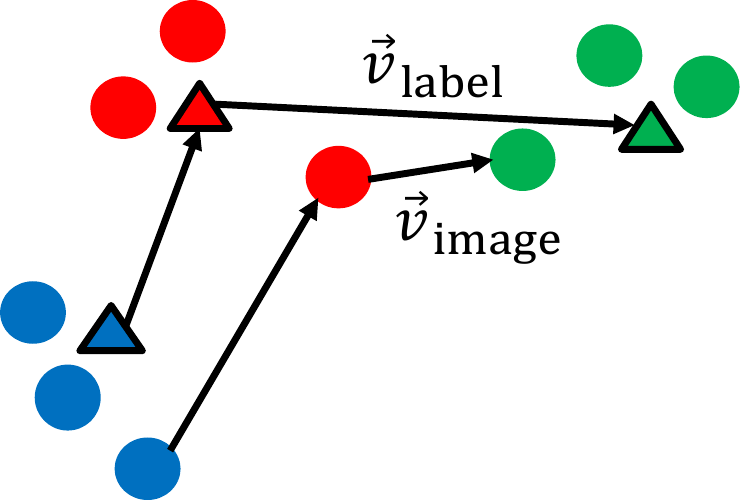} & 
\includegraphics[width = 0.4\linewidth]{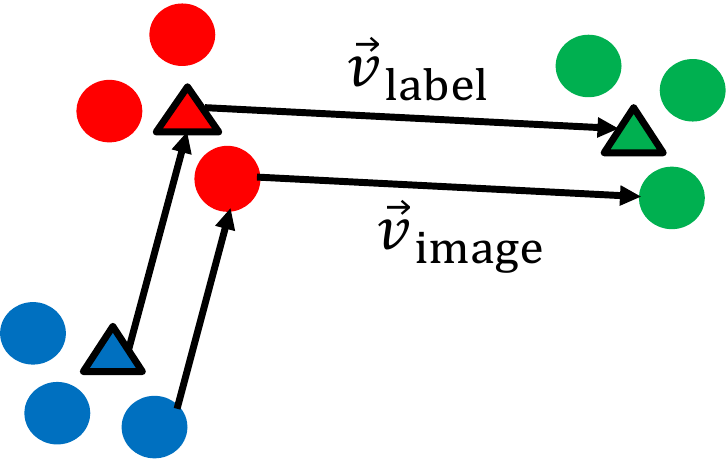} \\
(c) w/o difference constraints & 
(d) w/ difference constraints 
\end{tabular}
\end{center}
\caption{
Illustration of the proposed constraints. Triangles represent label embeddings and circles represent image embeddings. 
The visual categories are color-coded. 
Discriminative constraints encourage small distance for image instances of the same class and large distance otherwise. 
Difference constraints align the difference vectors of image embeddings with those of label embeddings. 
}
\vspace{-2mm}
\label{figure: constraints}
\end{figure}

\subsubsection{Baseline Model}
\label{section: baseline}

Our baseline model aims at projecting image instances to nearby positions with the corresponding labels in the semantic space.
We use the ranking loss (\ref{eq: rank}) to learn such an embedding,
%
\begin{equation}
\label{eq: rank}
\small
\begin{aligned}
&L_\mathrm{R}(\bx_i, y_i) \\
&=\sum_{y \neq y_i} \max (0, m + d(f(\bx_i), s(y_i)) - d(f(\bx_i), s(y))),
\end{aligned}
\end{equation}
where we measure the similarity between image and label embeddings based on the cosine distance, \ie, $d(f(\bx), s(y)) = 1-f(\bx) \cdot s(y)$.
%
%

\subsubsection{Discriminative Constraints}
\label{section: discriminative}

The ranking loss (\ref{eq: rank}) optimizes only the distance between the image and label embeddings. 
However, it does not capture the relationships \emph{among} image embeddings. This may lead to a small margin between images of different classes (see Figure~\ref{figure: constraints}(a) for an illustrative example) and limit the discrimination ability of the learned image embeddings.
We propose to explicitly model the intra-class and inter-class relationships of image embeddings. 
Specifically, we apply two alternative discriminative constraints to improve the baseline embedding model. 
First, the contrastive loss (\ref{eq: contrastive}) encourages small distance of two images from the same class and large distance otherwise. 
Second, the triplet loss (\ref{eq: triplet}) enforces distance between a reference image and an image from the same class to be smaller than that between the reference image and an image from a different class.
With discriminative constraints, image embeddings of the same class are more compact and those of different classes are easier to distinguish, as shown in Figure~\ref{figure: constraints}(b).
The contrastive loss function is of the form:
\begin{equation}
\label{eq: contrastive}
\small
\begin{aligned}
&L_\mathrm{C}(\bx_i, y_i, \bx_j, y_j) \\
&= \mathbf{1}_{(y_i = y_j)} d\left(f(\bx_i), f(\bx_j)\right) \\
&+ \mathbf{1}_{(y_i \neq y_j)} \max \left(m - d(f(\bx_i), f(\bx_j)), 0\right),
\end{aligned}
\end{equation}
where $\mathbf{1}_{(\cdot)}$ is the indicator function. 
The triplet loss function is defined as follows:
\begin{equation}
\label{eq: triplet}
\small
\begin{aligned}
&L_\mathrm{T}(\bx_i, y_i, \bx_j, y_j, \bx_k, y_k) \\
&= \max (0, m+d\left(f(\bx_i), f(\bx_j)\right)-d\left(f(\bx_i), f(\bx_k)\right)),
\end{aligned}
\end{equation}
where $\bx_i$ denotes the reference image, $\bx_j$ is an image from the same class ($y_i = y_j$), and $\bx_k$ from a different class ($y_i \neq y_k$).

\subsubsection{Difference Constraints}
\label{section: difference}

While discriminative constraints enforce image instances of different categories to be distant, there are no constraints on \emph{how} the two instances should be pulled apart.
To regularize the learning of visual-semantic embedding model, we propose to align the difference vectors of image pairs with those of label pairs.
The difference constraints are capable of preserving the semantic relationships among the label embeddings.
Figure~\ref{figure: constraints}(c) and (d) illustrate the effect of learning image embeddings with and without using the difference constraints. 

We formulate the difference constraints as follows:
\begin{equation}
\label{eq: difference}
\small
\begin{aligned}
&L_\mathrm{D}(\bx_i, y_i, \bx_j, y_j) \\
&= {\lVert (f(\bx_i) - f(\bx_j)) - (s(y_i) - s(y_j)) \rVert}^2_2,
\end{aligned}
\end{equation}
where $f(\bx_i) - f(\bx_j)$ indicates the difference vector of the two image embeddings and $s(y_i) - s(y_j)$ indicates the difference vector for their corresponding label embeddings.

\subsubsection{Objective Function}
\label{section: full}

We combine the embedding baseline and two extra structured constraints in a unified learning formulation. 
The overall training objective function can be represented as:
\begin{equation}
\label{eq: full-c}
\small
\begin{aligned}
\min_\Theta &\frac{w}{2}||\Theta||^2 + \lambda_1 \sum_i L_\mathrm{R}(\bx_i, y_i) \\
&+ \lambda_2 \sum_{i, j} L_\mathrm{C}(\bx_i, y_i, \bx_j, y_j) \\
&+ \lambda_3 \sum_{i, j} L_\mathrm{D}(\bx_i, y_i, \bx_j, y_j),
\end{aligned}
\end{equation}
or
\begin{equation}
\label{eq: full-t}
\small
\begin{aligned}
\min_\Theta &\frac{w}{2}||\Theta||^2+ \lambda_1 \sum_i L_\mathrm{R}(\bx_i, y_i) \\
&+ \lambda_2 \sum_{i, j, k} L_\mathrm{T}(\bx_i, y_i, \bx_j, y_j, \bx_k, y_k) \\
&+ \lambda_3 \sum_{i, j} L_\mathrm{D}(\bx_i, y_i, \bx_j, y_j),
\end{aligned}
\end{equation}
where $\Theta$ is the parameters of the image embedding function, \ie, the network weights, and $w = 0.0005$ represents the constant weight decay. The weights $\lambda_1, \lambda_2, \lambda_3$ balance these constraints. 

\subsection{Implementation Details}
\label{section: details}

\begin{figure}[!t]
\small
\begin{center}
\begin{tabular}{@{}c@{}}
\includegraphics[width = 0.9\linewidth]{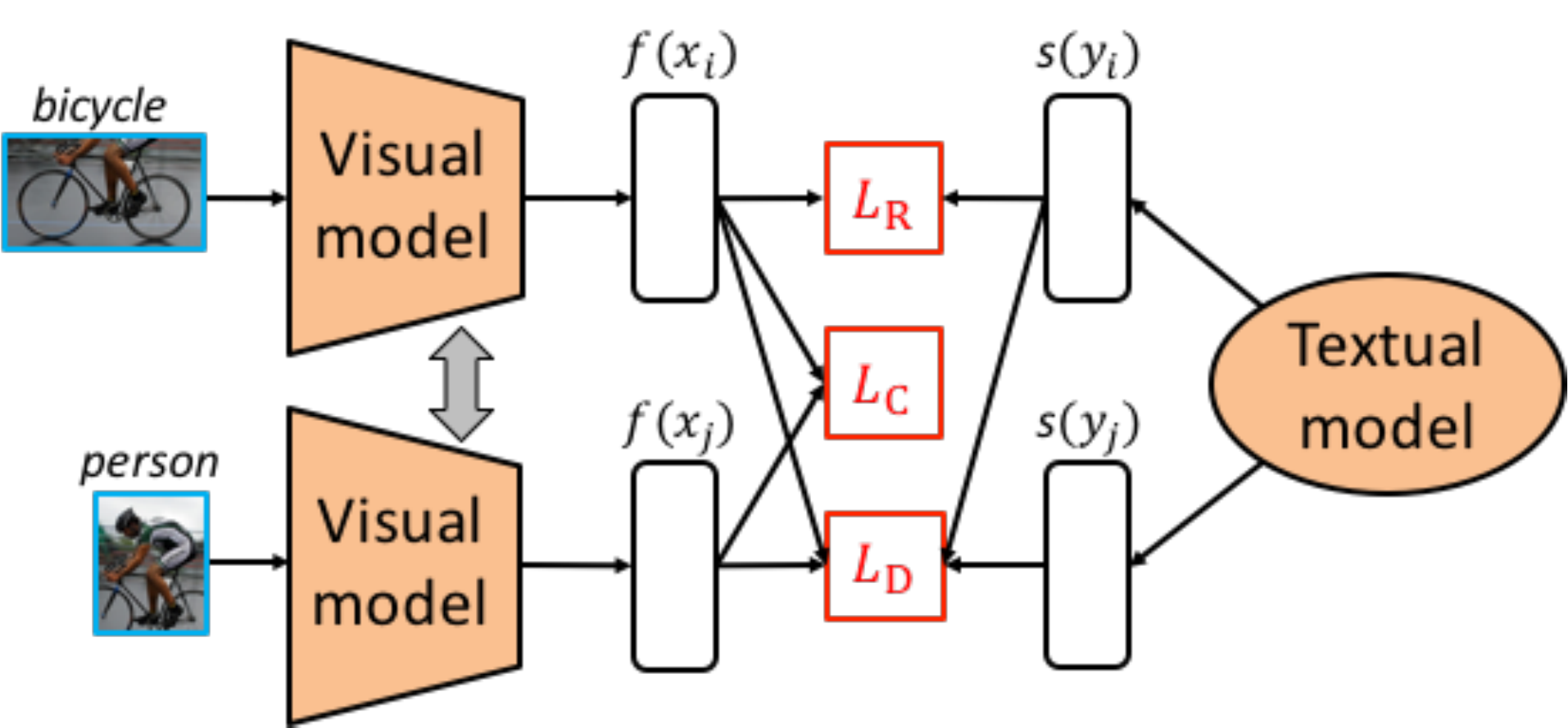} \\
(a) Two-branch network \vspace{3mm} \\
\includegraphics[width = 0.9\linewidth]{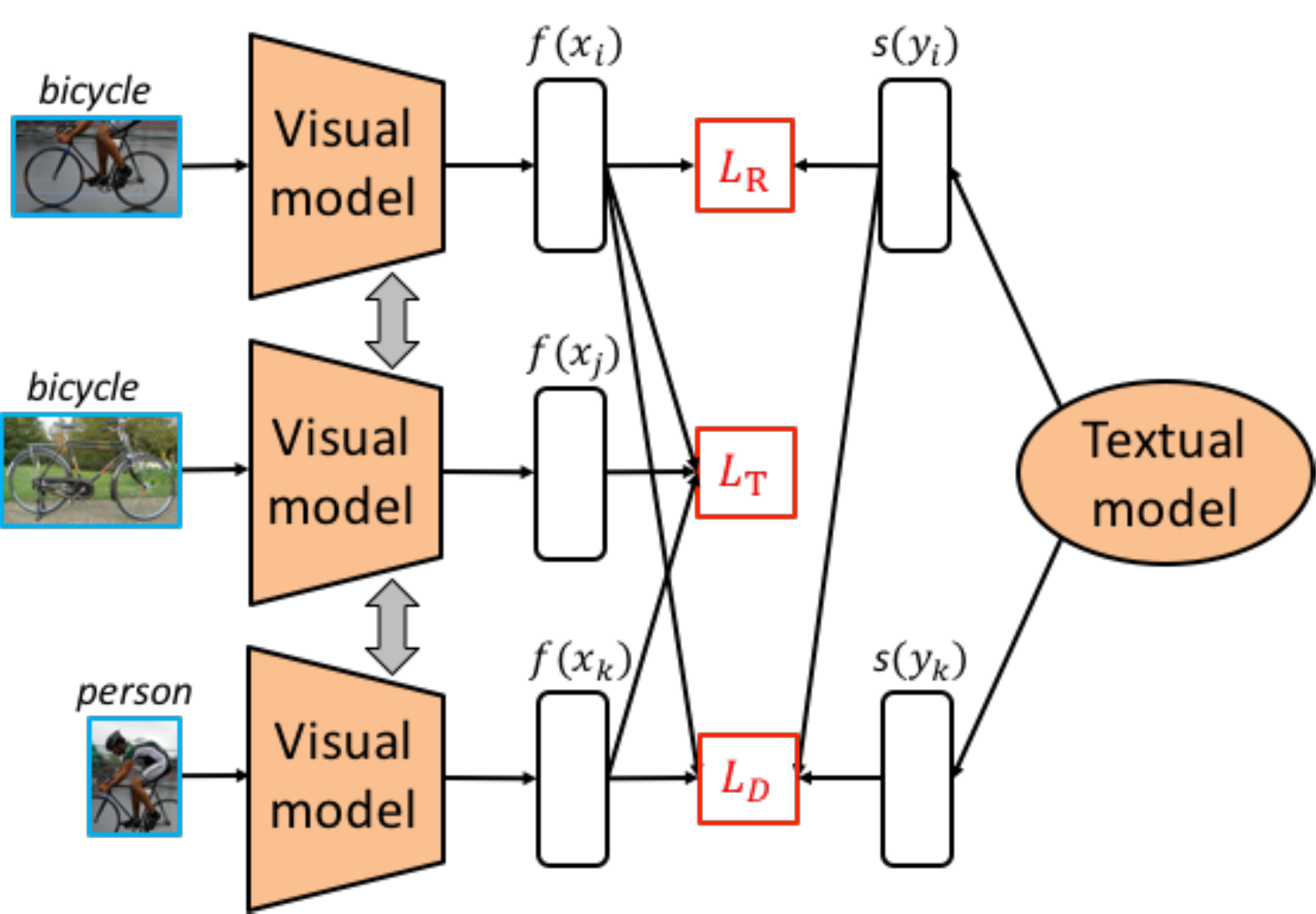} \\
(b) Three-branch network \\
\end{tabular}
\end{center}
\caption{Overview of our network architectures for learning structured embeddings. (a) The two-branch network takes pairs of training instances as input. (b) The three-branch network takes triplets of training instances as input. 
The base networks share the same architecture and parameter weights. See text for more details of loss functions: $L_\mathrm{R}$ (\ref{eq: rank}), $L_\mathrm{C}$ (\ref{eq: contrastive}), $L_\mathrm{T}$ (\ref{eq: triplet}), $L_\mathrm{D}$ (\ref{eq: difference}). 
}
\vspace{-2mm}
\label{figure: network}
\end{figure}

As shown in Figure~\ref{figure: network}, we build a two-branch (for contrastive loss) or three-branch (for triplet loss) network to learn image embeddings. We use the AlexNet~\cite{krizhevsky2012imagenet} (for CIFAR-10/100) and GoogLeNet~\cite{szegedy2014going} (for the other datasets) pre-trained on the ImageNet 2012 classification task as our base network architectures.
Each base network shares the same architecture and parameter weights. 
We add a linear transformation layer and a normalization layer with randomly initialized parameters on top of the output of each base network.
The new transformation layer projects image features to the 300D embeddings.
Both image and label embeddings are normalized to unit norm. 
Since semantic labels may indicate scenes/events/objects in the image, we use the general region proposal method~\cite{krahenbuhl2014geodesic} to collect candidate regions. 
%
%
Around 1,000 initial region proposals are generated for each image.
%
%
We then constrain the region width/height to be at least 0.3 of the image width/height and the aspect ratio to be within the range $[0.25,~4]$.

We use the Caffe toolbox~\cite{jia2014caffe} to train CNNs with a Tesla K40 GPU. 
Since optimizing over all pairs or triplets of instances is computationally infeasible, we randomly sample image instances for training. 
For the two-branch network, we use equal amounts of image pairs from the same and different classes in a batch. 
For the three-branch network, we use 20\% image instances from the same class of the reference image and the rest from different classes in a batch.
We set the initial learning rate to 0.001 with a step decay policy and the momentum to 0.9. 
We set the margins $m=0.1$ for the ranking baseline (\ref{eq: rank}) and $m=1.0$ for both the contrastive (\ref{eq: contrastive}) and triplet (\ref{eq: triplet}) losses in our experiments. 
The optimal balance weights $\lambda_1, \lambda_2, \lambda_3$ may be different for different datasets. 
%
%
We include the detailed analysis of hyper-parameters in the supplementary material.
The source code and pre-trained models will be made publicly available.

\section{Experimental Results}
\label{section: experiments}

In this section, we present extensive experimental results to demonstrate the effectiveness of our structured embeddings for visual recognition, including single-label classification, multi-label classification, and zero-shot recognition. We also analyze the contributions of individual components of our approach.

\subsection{Datasets}
\label{section: datasets}

\paragraph{Single-label classification.} 
We first use the CIFAR-10/100 datasets~\cite{krizhevsky2009learning} to validate the effectiveness of the proposed constraints. 
The CIFAR-10 dataset is composed of 10 categories of images with 50,000 training images, and 10,000 testing images. The CIFAR-100 dataset consists of 100 categories. There are 600 images for each category (500 for training and 100 for testing).
We evaluate our method for image classification on the large-scale ImageNet 2012 dataset with 1,000 labels~\cite{deng2009imagenet}. We use the default \emph{train} set to train the embedding model with structured constraints and test on the \emph{validation} set. 

\vspace{-3mm}
\paragraph{Multi-label classification.}
For multi-label image classification, we use the NUS-WIDE dataset~\cite{chua2009nus}.
This dataset contains 209,347 images and 81 semantic concepts.
We use the train/test split as in~\cite{gong2013deep} and use a subset of 150,000 images for training and the rest of the images for testing. 

\vspace{-3mm}
\paragraph{Zero-shot recognition.}
For zero-shot recognition, we use the aPascal \& aYahoo (aP\&Y)~\cite{farhadi2009describing} and ImageNet 2010~\cite{deng2009imagenet} datasets. 
We follow the standard split on the aP\&Y dataset. The Pascal set serves as training data, and the Yahoo set as test data for evaluation. 
%
%
%
%
For the ImageNet 2010 dataset, we use the 800/200 split of the 1,000 classes as in~\cite{frome2013devise}: training the embedding model using 800 classes, and inferring image labels using the rest 200 classes.

\subsection{Single-Label Classification}
\label{section: single-label}

\paragraph{Evaluations on CIFAR-10/100.}
We train a multi-class linear SVM based on the learned 300D embedding features for classification.
Table~\ref{table: cifar} shows the mean classification accuracy using the models trained with different constraints.
Combined with discriminative constraints only, we achieve higher accuracy than the baseline model, \eg, a 5.8\% gain on CIFAR-10 (86.9\% vs. 81.1\%). 
The results demonstrate the effectiveness of discriminative constraints for learning discriminative image embeddings to distinguish visual categories. 
Combined with difference constraints only, we obtain 2.9\% improvement on CIFAR-10 but slightly worse results on CIFAR-100. 
The intra- and inter-class relationships among different classes are important for image classification. 
Without explicitly modeling such relationships, difference constraints are not sufficient to distinguish visually similar categories.
Combining both constraints, we further improve the classification accuracy on both datasets. 
The full model trained with triplet and difference losses outperforms the ranking baseline by 6.7\% on CIFAR-10 and 1.9\% on CIFAR-100. 
The results validate that our two complementary structured constraints help improve the embedding model for visual recognition.
We also use the nearest neighbor classifier where the class label is inferred with the smallest distance between the image and all the candidate label embeddings. 
We obtain similar performance with SVM (87.0\% on CIFAR-10 and 63.8\% on CIFAR-100 using the model trained with triplet and different losses).

\vspace{-3mm}
\paragraph{Evaluations on ImageNet.}
Our model trained with the contrastive and difference losses achieves 60.7\% mean classification accuracy (\ie, flat hit@1), showing a 10.6\% relative improvement over the DeViSE method~\cite{frome2013devise}. 
The results demonstrate the effectiveness of our structured embeddings for large-scale image recognition.

\begin{table}[t]
\caption{Mean classification accuracy (\%) on the CIFAR-10/100 datasets using different constraints. {\bf\color{red}{Red color}} and \underline{\color{blue}{blue color}} indicate the best and second best performing algorithms, respectively.}
\label{table: cifar}
\vspace{1mm}
\centering
\small
\resizebox{\linewidth}{!}{
\begin{tabular}{c c c c |c c}\toprule
\multicolumn{4}{c|}{Constraints} & \multirow{2}{*}{CIFAR-10} & \multirow{2}{*}{CIFAR-100} \\
Rank & Contrastive & Triplet & Difference & & \\
\midrule
$\surd$ & & & & 81.1 & 62.4 \\
& $\surd$ & & & 73.5 & 42.2 \\
& & $\surd$ & & 84.2 & 53.8 \\
& & & $\surd$ & 80.0 & 58.2 \\
$\surd$ & $\surd$ & & & 83.1 & 62.7 \\
$\surd$ & & $\surd$ & & \underline{\color{blue}86.9} & 62.5 \\
$\surd$ & & & $\surd$ & 84.0 & 61.0 \\
$\surd$ & $\surd$ & & $\surd$ & 84.6 & \underline{\color{blue}63.2} \\
$\surd$ & & $\surd$ & $\surd$ & \bf\color{red}87.8 & \bf\color{red}64.3 \\
\bottomrule
\end{tabular}
}
\vspace{-2mm}
\end{table}

\begin{table}[t]
\caption{Comparisons of visual-semantic models for image classification on the ImageNet 2012 datasets with 1,000 classes in terms of flat hit@k metrics.}
\label{table: imagenet}
\vspace{1mm}
\centering
\small
\resizebox{\linewidth}{!}{
\begin{tabular}{l c c c c } \toprule
Models & $k=1$ & $k=2$ & $k=5$ & $k=10$ \\
\midrule
Norouzi~\etal~\cite{norouzi2013zero} & 54.3 & 61.9 & 68.0 & 71.6 \\
Frome~\etal~\cite{frome2013devise} & 54.9 & 66.9 & \underline{\color{blue}78.4} & \underline{\color{blue}85.0} \\
Rank + Contrastive + Difference & \bf\color{red}60.7 & \bf\color{red}72.1 & \bf\color{red}82.6 & \bf\color{red}87.9 \\
Rank + Triplet + Difference & \underline{\color{blue}57.3} & \underline{\color{blue}67.7} & 77.8 & 83.3 \\
\bottomrule
\end{tabular}
}
\vspace{-2mm}
\end{table}

\subsection{Multi-Label Classification}
\label{section: multi-label}

%
We train one-versus-all SVM classifiers for multi-class classification. At test time, we extract one feature vector for each entire image and compute the prediction scores of each class by those SVM classifiers. 
We also use the setting with region proposals for inference, but find that the predicted regions are not accurate for multi-label classification.
%
%
Following the same evaluation protocols~\cite{gong2013deep,ren2015multi,Wang-CVPR-2016}, we generate $k$ (\eg, $k=3$) highest ranked labels for each test image and then compute the per-class precision (C-P), per-class recall (C-R), per-class F1 (C-F1), overall precision (O-P), overall recall (O-R), and overall F1 (O-F1) scores. 
The mean average precision (mAP)@N is also used in the recent work~\cite{Wang-CVPR-2016}.
However, we note that these scores are computed based on a \emph{fixed} number of predicted labels for each image. 
Such metrics are not sufficiently accurate as each image may have a different number of labels. 
In light of this, we also report the widely used mean average precision (mAP)~\cite{everingham2010pascal} measure.

\begin{table*}[t]
\caption{Comparisons of image classification performance on the NUS-WIDE dataset. The precision/recall/F1 scores are computed with $k=3$ predicted labels per image. The ranking loss is enabled in all of our embedding models.}
\label{table: nuswide}
\small
\vspace{1mm}
\centering
\begin{tabular}{l|ccccccccccc} \toprule
Methods & & & & C-P & C-R & C-F1 & O-P & O-R & O-F1 & mAP@10 & mAP \\
\midrule
Li~\etal~\cite{li2015distributed} & & & & - & - & - & - & - & 21.3 & - & - \\
Liu~\etal~\cite{liu2010unified} & & & & - & - & - & 35.0 & 37.0 & 36.0 & - & - \\
Chua~\etal~\cite{chua2009nus} & & & & 32.6 & 19.3 & 24.3 & 42.9 & 53.4 & 47.6 & - & - \\
Gong~\etal~\cite{gong2013deep} & & & & 31.7 & 35.6 & 33.5 & 48.6 & 60.5 & 53.9 & - & - \\
Wang~\etal~\cite{Wang-CVPR-2016} & & & & \bf\color{red}40.5 & 30.4 & 34.7 & 49.9 & 61.7 & 55.2 & 56.1 & - \\
Wu~\etal~\cite{wu2015cross} & & & & - & - & - & - & - & - & 40.3 & - \\
Ren~\etal~\cite{ren2015multi} & & & & 37.7 & 40.2 & 38.9 & 52.2 & \bf\color{red}65.0 & \underline{\color{blue}57.9} & - & - \\
\midrule
\multirow{2}{*}{W/O region mining} & \multicolumn{3}{c}{W/ random regions} & 29.1 & 3.7 & 6.6 & 29.1 & 35.6 & 32.0 & 35.6 & 3.5 \\
& \multicolumn{3}{c}{W/ entire images} & 33.0 & 37.6 & 35.2 & 50.8 & 62.0 & 55.8 & 73.7 & 37.0 \\
\midrule
\multirow{8}{*}{W/ region mining} & Contrastive & Triplet & Difference & & & & & & & & \\
& & & & 35.7 & 42.9 & 39.0 & 52.3 & 63.9 & 57.5 & 76.4 & 42.4 \\
& $\surd$ & & & 35.8 & 43.3 & 39.2 & 52.2 & 63.8 & 57.4 & 76.3 & 42.5 \\
& & $\surd$ & & 38.0 & 41.2 & 39.5 & 52.3 & 63.8 & 57.5 & 76.6 & 42.2 \\
& & & $\surd$ & 36.3 & \underline{\color{blue}44.0} & \underline{\color{blue}39.8} & 52.4 & 64.0 & 57.6 & 76.6 & 43.2 \\
& $\surd$ & & $\surd$ & \underline{\color{blue}38.9} & \bf\color{red}44.6 & \bf\color{red}41.6 & \bf\color{red}52.9 & \underline{\color{blue}64.6} & \bf\color{red}58.2 & \bf\color{red}77.6 & \bf\color{red}46.1 \\
& & $\surd$ & $\surd$ & 38.0 & 40.9 & 39.4 & \underline{\color{blue}52.7} & 64.3 & \underline{\color{blue}57.9} & \underline{\color{blue}77.2} & \underline{\color{blue}43.9} \\
\bottomrule 
\end{tabular}
\vspace{-2mm}
\end{table*}

\vspace{-3mm}
\paragraph{Comparisons to the state-of-the-art methods.}
We compare the proposed visual-semantic embedding approach with the state-of-the-art methods for multi-label image classification, including metric learning~\cite{li2015distributed}, multi-edge graph~\cite{liu2010unified}, KNN~\cite{chua2009nus}, cross-modal ranking~\cite{wu2015cross}, WARP~\cite{gong2013deep}, MIE~\cite{ren2015multi} and CNN-RNN~\cite{Wang-CVPR-2016} methods.
Table~\ref{table: nuswide} shows quantitative results on the NUS-WIDE dataset.
Overall, compared to the methods without learning visual-semantic embeddings~\cite{li2015distributed,liu2010unified,chua2009nus,gong2013deep,Wang-CVPR-2016}, our embedding baseline achieves notable improvements in terms of different metrics, \eg, a 10\% gain in O-F1 over the KNN baseline~\cite{chua2009nus}. 
We attribute the performance improvement to the rich semantic relationships among word embeddings, which help better recognize visual categories.
Our full model trained with contrastive and difference losses performs favorably against the previous visual-semantic methods~\cite{ren2015multi,wu2015cross}, \eg, a 2.7\% gain over~\cite{ren2015multi} in C-F1 and 37.7\% gain over~\cite{wu2015cross} in mAP@10.
The results demonstrate the effectiveness of our structured constraints for learning discriminative embedding model.
In addition, our full model outperforms the state-of-the-art method~\cite{Wang-CVPR-2016} by 20\% in terms of mAP@10.
The results suggest that we get accurate ranked lists in the top 10 predictions for each test image.


\begin{figure}[t]
\centering
\includegraphics[width=\linewidth]{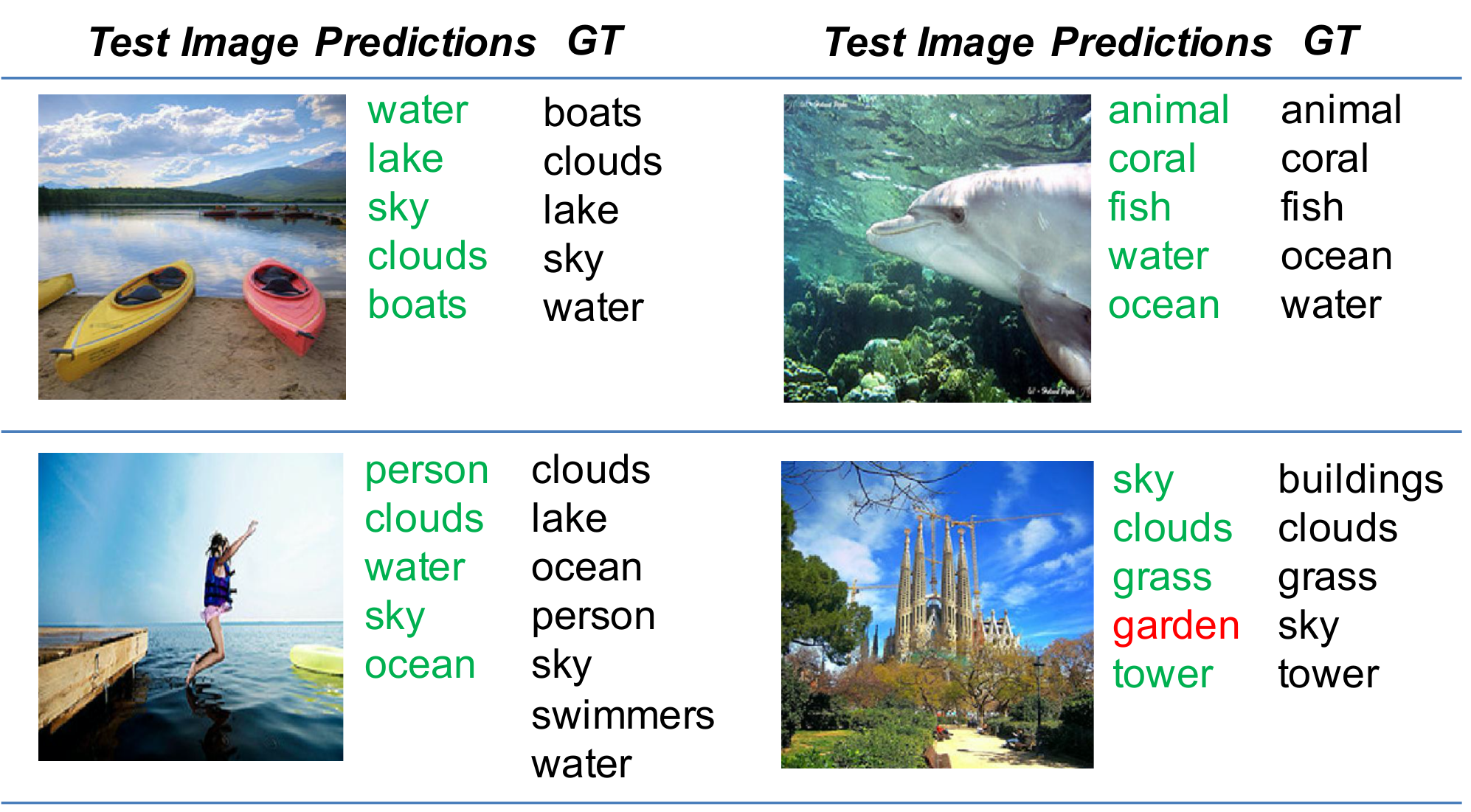}
\caption{Examples of classification results obtained by our embedding model on the NUS-WIDE dataset. We show top 5 predicted labels for each test image. Green texts indicate correct predictions and red texts indicate wrong predictions. Note that the predictions are ranked and ground-truth labels are not ranked.}
\label{figure: predictions}
\end{figure}

\vspace{-3mm}
\paragraph{Contributions from individual components.}
We also show the relative contributions of the mined regions and the proposed constraints in Table~\ref{table: nuswide}. 
Our embedding baseline with mined regions achieves 42.4\% mAP, significantly outperforming those with random regions or entire images. 
The performance gain comes from the multi-label training procedure. 
%
%
Compared to the embedding baseline with only the ranking loss, our model learned with both constraints achieves higher performance in terms of all the metrics (\eg, 3.7\% improvement in mAP). 
This demonstrates that our structured constraints help improve the discrimination ability of image embeddings. 
We observe that there is no significant improvement if we learn the embedding model with discriminative or difference constraints individually. 
The results show again that the two constraints are complementary for learning visual-semantic embeddings.

\vspace{-3mm}
\paragraph{Qualitative results.}
%
Figure~\ref{figure: predictions} shows examples of prediction results on the test set. 
Our embedding model makes correct predictions for different semantic concepts, \eg, objects and scenes. 
The lower right image in Figure~\ref{figure: predictions} shows a typical failure case. 
Our model incorrectly predicts \emph{garden} that is visually and semantically similar to the ground-truth label of \emph{grass}. 
While such predictions are incorrect, they are semantically plausible.
We refer the readers to the supplementary material for more results.

\subsection{Zero-Shot Recognition}
\label{section: zero-shot}

One of the critical applications of visual-semantic embedding is zero-shot recognition. 
%
For each test image, we first extract the 300D image embedding based on the learned model. 
We then infer the class label using nearest neighbor search.

\vspace{-3mm}
\paragraph{Comparisons to the state-of-the-art methods.} 
We compare the proposed approach with the state-of-the-art methods\footnote{The results of \cite{norouzi2013zero,frome2013devise,akata2015evaluation} are from~\cite{XianCVPR2017} with the ResNet features.} for zero-shot recognition on the aP\&Y dataset in Table~\ref{table: apy}. 
With word embedding only, we achieve higher accuracy over the existing word embedding based methods, \eg, outperforming DeViSE~\cite{frome2013devise} by 6.7\%. 
The performance gain can be explained by our structured constraints. 
To combine word and attribute embeddings, we train an additional embedding model with the ranking loss based on the normalized 64D attribute vectors. The word and attribute embeddings are then concatenated for classification using the nearest neighbor search.
With the combined embeddings, we obtain competitive results with the state-of-the-art algorithms.
Note that our method does not rely on the ground-truth object bounding boxes for training the embedding model. 
The recent work~\cite{zhang2016zero2} improves the zero-shot recognition performance by adapting the learned similarity functions using the test data. 
In contrast, we leverage the nearest neighbor classifier for each test image individually and do not impose additional assumptions on the test data.

\begin{table}[t]
\caption{Zero-shot image classification accuracy (\%) on the aP\&Y dataset. The ranking loss is enabled in all of our embedding models.}
\label{table: apy}
\vspace{+1mm}
\centering
\small
\resizebox{\linewidth}{!}{
\begin{tabular}{cc|cccc} \toprule
\multicolumn{2}{c|}{Semantic sources} & \multicolumn{3}{l}{\multirow{2}{*}{Methods}} & \multirow{2}{*}{Accuracy} \\
Words & Attributes & & & & \\
\midrule
& $\surd$ & \multicolumn{3}{l}{Romera-Paredes~\etal~\cite{romera2015embarrassingly}} & 27.3 \\
& $\surd$ & \multicolumn{3}{l}{Lampert~\etal~\cite{lampert2014attribute}} & 38.2 \\
& $\surd$ & \multicolumn{3}{l}{Zhang~\etal~\cite{zhang2016zero1}} & 50.4 \\
& $\surd$ & \multicolumn{3}{l}{Bucher~\etal~\cite{bucher2016improving}} & \underline{\color{blue}53.2} \\
$\surd$ & & \multicolumn{3}{l}{Norouzi~\etal~\cite{norouzi2013zero}} & 25.9 \\
$\surd$ & & \multicolumn{3}{l}{Frome~\etal~\cite{frome2013devise}} & 35.4 \\ 
$\surd$ & $\surd$ & \multicolumn{3}{l}{Akata~\etal~\cite{akata2015evaluation}} & 32.0 \\ 
\midrule
\multicolumn{2}{c|}{} & Contrastive & Triplet & Difference & \\
$\surd$ & & & & & 33.7 \\
$\surd$ & & $\surd$ & & & 40.7 \\
$\surd$ & & & $\surd$ & & 34.3 \\
$\surd$ & & & & $\surd$ & 37.4 \\
$\surd$ & & $\surd$ & & $\surd$ & 42.1 \\
$\surd$ & & & $\surd$ & $\surd$ & 40.2 \\
$\surd$ & $\surd$ & & & & 47.5 \\
$\surd$ & $\surd$ & $\surd$ & & & 48.0 \\
$\surd$ & $\surd$ & & $\surd$ & & 47.8 \\
$\surd$ & $\surd$ & & & $\surd$ & 49.0 \\
$\surd$ & $\surd$ & $\surd$ & & $\surd$ & \bf\color{red}54.7 \\
$\surd$ & $\surd$ & & $\surd$ & $\surd$ & 51.1 \\
\bottomrule
\end{tabular}
}
\end{table}

\begin{table}[t]
\caption{Comparisons of zero-shot recognition on the ImageNet 2010 datasets in terms of flat hit@5 accuracy (\%).}
\label{table: imagenet-zero}
\vspace{1mm}
\centering
\small
\resizebox{\linewidth}{!}{
\begin{tabular}{l c c c c } \toprule
Models & 200 labels &1,000 labels \\
\midrule
Norouzi~\etal~\cite{norouzi2013zero} & 28.5 & - \\
Frome~\etal~\cite{frome2013devise} & 31.8 & 9.0 \\
Rohrbach~\etal~\cite{rohrbach2011evaluating} & 34.8 & - \\
Mensink~\etal~\cite{mensink2012metric} & 35.7 & 1.9 \\
Fu~\etal~\cite{fu2015zero} & 41.0 & - \\
Mukherjee~\etal~\cite{mukherjee2016gaussian} & 45.7 & - \\
Huang~\etal~\cite{huang2016local} & \bf\color{red}48.2 & - \\
\midrule
Rank + Contrastive + Difference & \underline{\color{blue}46.2} & \bf\color{red}12.4 \\
Rank + Triplet + Difference & 45.0 & \underline{\color{blue}11.3} \\
\bottomrule 
\end{tabular}
}
\vspace{-2mm}
\end{table}

\vspace{-3mm}
\paragraph{Contributions from individual components.}
We also show the effect of the proposed constraints in Table~\ref{table: apy}. 
Our full models consistently outperform the embedding baseline model with the ranking loss by 8.4\% with word embedding only and 7.2\% with combined embeddings. 
%
%
Our models with either discriminative or difference constraints also improve the embedding baseline. 
Combing the two constraints brings further improvement, which shows the complementary nature of the two structured constraints.

\begin{figure}[t]
\small
\begin{center}
\begin{tabular}{@{}cc@{}}
\includegraphics[width = 0.52\linewidth]{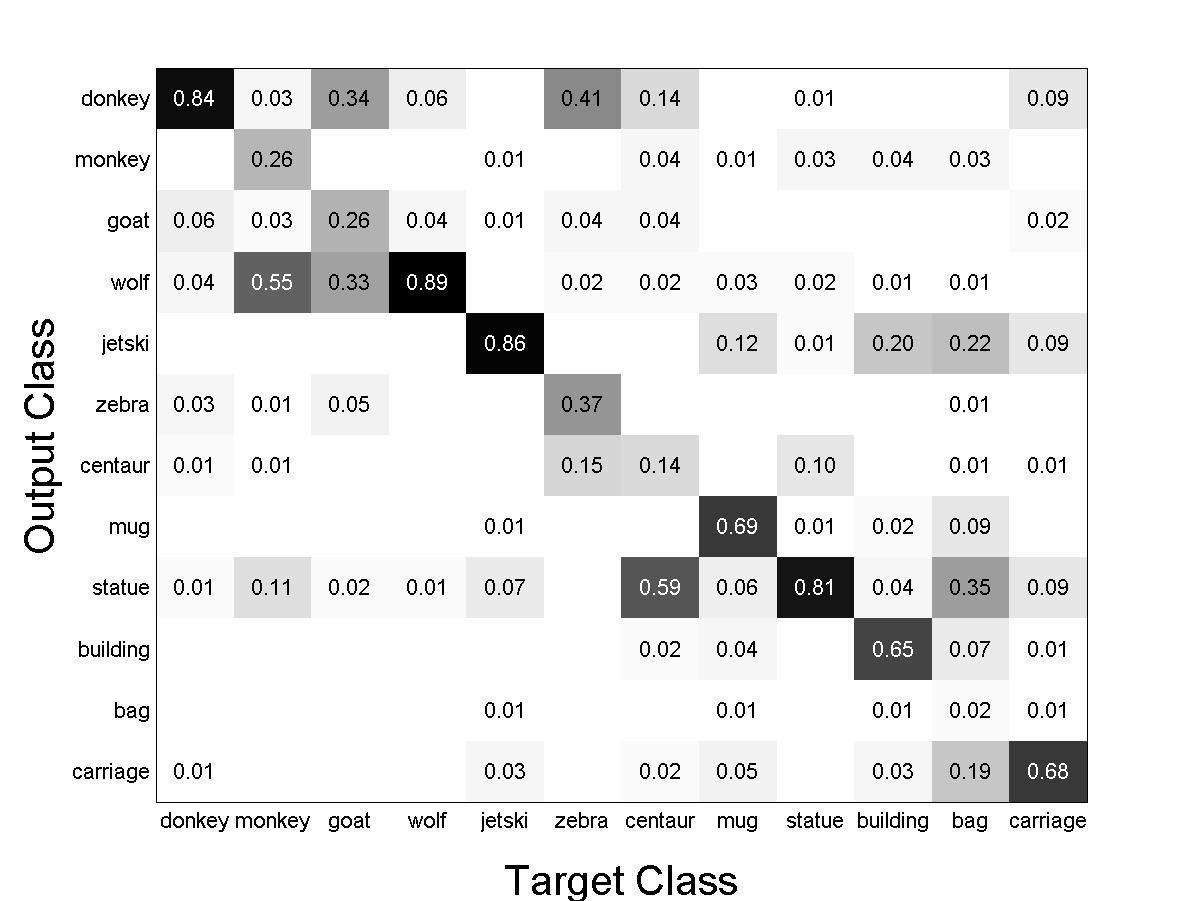} & 
\includegraphics[width = 0.42\linewidth]{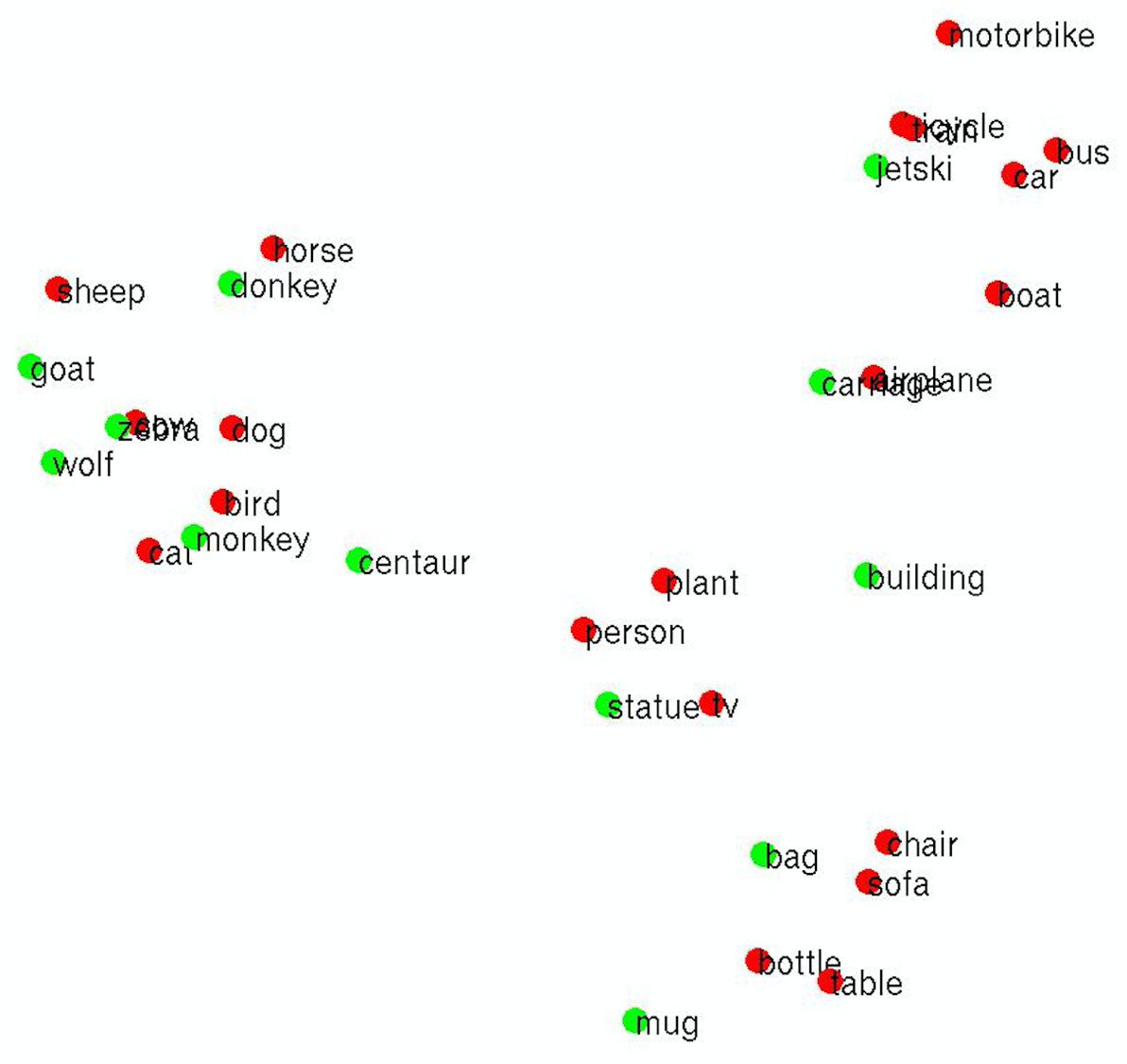} \\
(a) Confusion matrix & 
(b) Label embeddings \\
\end{tabular}
\end{center}
\caption{Error analysis of zero-shot recognition by our method on the aP\&Y dataset. (a) Confusion matrix computed based on the per-class prediction results using our full model. Most errors come from semantically similar classes. (b) The label embeddings of both seen classes from Pascal (red) and unseen classes from Yahoo (green). The object classes with similar semantics are close in the embedding space. }
\label{figure: error}
\end{figure}

\vspace{-3mm}
\paragraph{Error analysis.}
We analyze per-class recognition results of our method with the confusion matrix in Figure~\ref{figure: error}(a). The majority of errors comes from confusion with semantically similar categories, \eg, \emph{zebra} and \emph{donkey}. This is because words with similar semantics are embedded at close positions in the space, as shown in Figure~\ref{figure: error}(b).

\vspace{-3mm}
\paragraph{Evaluations on ImageNet.}
We show the capacity of our visual-semantic embedding models for zero-shot recognition when there are large amounts of class labels. 
%
Table~\ref{table: imagenet-zero} shows performance comparisons in terms of flat hit@5 accuracy. 
Our models learned with the proposed structured constraints achieve comparative performance with the state-of-the-art methods. 
When classifying test data into the joint label space of both seen and unseen classes, we also achieve better accuracy (a 3.4\% gain over DeViSE~\cite{frome2013devise}). This indicates that our models have less bias toward training classes than the previous methods.


\section{Conclusions}

In this paper, we propose to incorporate two structured constraints for learning visual-semantic embeddings.
Discriminative constraints model the intra- and inter-class relationships and difference constraints serve as a regularizer to preserve the semantic relationships among word embeddings. 
Quantitative results show that the two constraints are complementary and crucial for improving visual recognition.
Our method is simple, flexible, and easily applicable to large amounts of categories since we do not rely on costly bounding box annotations. 
Experimental evaluations on multiple datasets including the large-scale ImageNet dataset demonstrate the effectiveness of our embedding model with structured constraints for image classification and zero-shot recognition. 
In the future work, we plan to jointly learn the visual and textual embeddings and explore additional applications, \eg, object localization using the visual-semantic embedding model.

\clearpage
{\footnotesize
\bibliographystyle{ieee}
\bibliography{semantic_embedding}
}

\end{document}